\documentclass[preprint,12pt]{elsarticle}

\usepackage{amssymb}
\usepackage{amsmath}
\usepackage{lineno}
\usepackage[colorlinks=true, citecolor=blue, linkcolor=blue, urlcolor=blue]{hyperref}

\journal{Computer Physics Communications}

\begin{document}

\begin{frontmatter}

\title{Integrating Physics and Data-Driven Approaches: An Explainable and  Uncertainty-Aware Hybrid Model for Wind Turbine Power Prediction}

\author[granada,cordoba]{Alfonso Gijón}\corref{corr}\ead{agijon@uco.es}
\author[poli]{Simone Eiraudo}
\author[granada]{Antonio Manjavacas}
\author[poli]{Daniele Salvatore Schiera}
\author[granada]{Miguel Molina-Solana}
\author[granada]{Juan Gómez-Romero}

\affiliation[granada]{organization={Department of Computer Science and Artificial Intelligence},
            addressline={Universidad de Granada},
            city={Granada},
            postcode={18071}, 
            state={Spain}}

\affiliation[poli]{organization={Energy Center Lab},
            addressline={Politecnico di Torino}, 
            city={Torino},
            postcode={10138}, 
            state={Italy}}

\affiliation[cordoba]{organization={Deparment of Informatics and Numerical Analysis},
            addressline={Universidad de Córdoba},
            city={Córdoba},
            postcode={14014}, 
            state={Spain}}

\cortext[corr]{Corresponding author.}

\begin{abstract}
The rapid growth of the wind energy sector underscores the urgent need to optimize turbine operations and ensure effective maintenance through early fault detection systems. While traditional empirical and physics-based models offer approximate predictions of power generation based on wind speed, they often fail to capture the complex, non-linear relationships between other input variables and the resulting power output. Data-driven machine learning methods present a promising avenue for improving wind turbine modeling by leveraging large datasets, enhancing prediction accuracy but often at the cost of interpretability. In this study, we propose a hybrid semi-parametric model that combines the strengths of both approaches, applied to a dataset from a wind farm with four turbines. The model integrates a physics-inspired submodel, providing a reasonable approximation of power generation, with a non-parametric submodel that predicts the residuals. This non-parametric submodel is trained on a broader range of variables to account for phenomena not captured by the physics-based component. The hybrid model achieves a 37\% improvement in prediction accuracy over the physics-based model. To enhance interpretability, SHAP values are used to analyze the influence of input features on the residual submodel's output. Additionally, prediction uncertainties are quantified using a conformalized quantile regression method. The combination of these techniques, alongside the physics grounding of the parametric submodel, provides a flexible, accurate, and reliable framework. Ultimately, this study opens the door for evaluating the impact of unmodeled variables on wind turbine power generation, offering a basis for potential optimization.
\end{abstract}


\end{frontmatter}


\section{Introduction}\label{sec:introduction}

The increasing adoption of renewable energy sources is crucial for addressing climate change and promoting a sustainable energy future \cite{IRENA}. Simultaneously, rapid advancements in sensor and storage technologies have enabled the collection of large volumes of data complemented by the emergence of flexible and powerful data-driven and machine learning techniques \cite{Wu2016,Yao2023}. Within this context, the development of accurate and robust wind turbine (WT) models is imperative for optimizing operations \cite{munteanu2008optimal} and enabling automated fault diagnosis \cite{Bindingsbo2023}.

Power prediction in WTs as a function of the weather conditions and the turbine state is crucial for wind energy management and power forecasting \cite{Mehrjoo2019}. Beyond its role in managing generated energy and integrating it into the electricity grid \cite{Watson1994}, power prediction, when combined with data analysis and monitoring techniques, aids in detecting and diagnosing powertrain component failures \cite{Benbouzid2021}. By comparing predicted power with real-time data, deviations and anomalies can be identified, potentially indicating component issues, thus enabling early intervention and effective fault management. Additionally, power prediction is vital for optimal pitch angle control and overall WT performance, as it provides real-time data on the power output. This information can be leveraged to adjust blade angles, maximize power generation, protect the WT, and ensure efficient system control \cite{APATA2020}.

The lack of precise and efficient physics-based models for forecasting power production in utility-scale wind farms has led to the adoption of data-driven approaches \cite{Pujana2023}. While machine learning approaches are traditionally regarded as black-box models, the development of novel architectures, such as physics-informed neural networks (PINNs), has enhanced their ability to model physical phenomena while ensuring accuracy and robustness \cite{Fernandez2023}. However, despite adhering to certain physical constraints, PINNs remain non-explainable and difficult to interpret. In contrast, hybrid semi-parametric models, which combine parametric physics-based and non-parametric data-driven methods, offer the advantage of high accuracy while maintaining the interpretability of the modeled functional relationships \cite{VONSTOSCH2014}. Explainability and uncertainty quantification are increasingly recognized as essential methods for gaining a deeper understanding of machine learning models \cite{Roscher2020,Abdar2021}. However, these best practices remain in the early stages of adoption within the wind energy sector, despite their significant potential for providing valuable scientific and engineering insights.

In this study, we design and validate a robust and accurate hybrid model for predicting the power generated by WTs at the `La Haute Borne' wind farm in France. The model combines a physics-based component to approximate the target output with a neural network that learns the residuals between the observed data and the physics-based approximation. An explainability analysis is conducted to assess the influence of input variables on the model's output, while conformalized predictions are used to quantify the uncertainties associated with the model.

The remainder of the paper is organized as follows: \autoref{sec:related-work} provides an overview of the related work and the main contributions of this study. \autoref{sec:methods} outlines the computational methods used to build the hybrid model, as well as those employed to study explainability and uncertainty quantification. \autoref{sec:case-study} delves into the analyzed real-world case study and offers practical implementation details. \autoref{sec:results} presents and discusses the experimental results, while the final conclusions and suggestions for future work are provided in \autoref{sec:conclusions}.

\section{Related work and main contributions}\label{sec:related-work}
Existing wind power prediction methods can generally be categorized into four types: physical methods, statistical methods, machine learning methods, and hybrid prediction methods \cite{CHEN2020}.

Detailed physical models leverage meteorological and geographical data to simulate the dynamic processes of wind \cite{CHEN2020}, offering high accuracy and strong physical interpretability. However, their computational efficiency is limited, as they must account for complex fluid dynamics and atmospheric effects \cite{Zehtabiyan2022}, making them impractical for large-scale data applications. In contrast, approximate empirical physical models, while useful for qualitative physical insights, often fail to accurately predict power production at utility-scale wind farms due to numerous simplifying assumptions and neglected physical phenomena \cite{Howland2019}. Several empirical formulas have been proposed to model the power coefficient; however, no consensus has been achieved regarding a model that offers sufficient flexibility and robustness \cite{Carpintero2020,Castillo2023}. Statistical methods, such as the simple moving average strategy, are commonly employed to forecast wind power using extensive historical data. However, the predictive performance of these models is constrained by their limited ability to capture nonlinear relationships \cite{ZHU2019}. Compared with physical methods, statistical methods are usually simpler and more suitable for small farms \cite{LIU2010}.

Data-driven machine learning models are extensively used for wind power forecasting due to their strong nonlinear fitting capabilities and self-learning abilities \cite{Clifton2013,HEINERMANN2016,SUN2020,DEMOLLI2019,MOSS2024,Dhungana2025}. The direct application of machine learning methods can yield favorable results in predicting the WT power curve and significantly contribute to turbine performance analysis. However, these methods often lack robustness and physical interpretability \cite{Wang2023}. For effective real-system optimization, the predictive model must accurately replicate observed data while adhering to the physical constraints of the system, even for atypical configurations.

Finally, hybrid methods combine physics-based with data-driven approaches to effectively address regression problems \cite{Stoffel2022}, having advantages such as a broader knowledge base, transparency in the modeling process, and cost-effective model development \cite{VONSTOSCH2014}. Multi-objective ensemble models have distinguished themselves among hybrid models by combining multiple data groups with different distributions and multiple learners to achieve superior predictive performance and generalization \cite{TASCIKARAOGLU2014,LIU2020}. However, there is still room for improvement through outlier correction, diversification of combination weights, or the use of residual data to refine prediction results and enhance model accuracy \cite{CHEN2022,YIN2022}. 

The non-parametric, data-driven component of hybrid regression models has a significant limitation: it lacks inherent functional explainability, and is therefore not directly interpretable. Furthermore, the underlying functional relationships may remain unclear, meaning causality between explanatory and dependent variables cannot be guaranteed. Recent advancements in explainability and uncertainty quantification techniques have significantly improved the interpretability and reliability of predictions made by data-driven models \cite{Linardatos2020,Abdar2021}. The application of these techniques to the WT power sector remains in its early stages and has been identified as a key research challenge in wind energy by the European Academy of Wind Energy (EAWE) \cite{Kuik2016}. Recent studies have explored Bayesian \cite{Aerts2023,Mclean2023} and Monte Carlo \cite{Richter2022} methods to approximate uncertainty in power curve model estimation, while others have begun employing explainability techniques to develop robust and transparent data-driven models \cite{CAKIROGLU2024,LIAO2024,LETZGUS2024}. To the best of our knowledge, no existing work has simultaneously incorporated both considerations—practices we deem essential for achieving trustworthy and robust power curve models.

In this work, we propose a hybrid model comprising a physics-based submodel for predicting the power generated by a dataset of four similar turbines and a data-driven, non-parametric submodel to learn the residuals between the physics-based submodel's output and the observed data. This approach aims to capture unmodeled physical phenomena and variations in the data that the physics-based submodel alone cannot fully account for. It is particularly advantageous when certain inputs are not reliably represented in the data or when integrating domain knowledge through a physical reference input is beneficial.
The main novelties and contributions of this paper can be summarized as follows:
\begin{itemize}
    \item A hybrid model is proposed for regression analysis of WT power generation as a function of weather conditions and internal turbine variables. This approach preserves the interpretability of the physics-based submodel while achieving high predictive accuracy.
    \item The proposed physics-based model operates with minimal assumptions regarding the functional form, deriving the generated power as a function of wind kinetic energy modulated by the power coefficient. The power coefficient is constrained by the Betz limit and is modeled using a neural network. This methodology offers significant flexibility while enabling the estimation of the power coefficient profile, which serves to characterize the aerodynamic performance of the WT.
    \item Prediction accuracy is improved by incorporating a non-parametric model that accounts for additional unmodeled input features to learn the residuals between the physics-based model's predictions and the observed data. An explainability analysis is conducted to evaluate the influence of input variables on the hybrid model's output. This analysis identifies the most significant features and provides insights into potential improvements for the physics-based model.
    \item To evaluate the reliability of the hybrid model's predictions, a conformalized quantile regression method is employed for uncertainty quantification, enabling the calculation of coverage and mean length of the uncertainty intervals. This approach enhances confidence in the model, which, to the best of our knowledge, is the first hybrid model in the wind energy sector to integrate both explainability and uncertainty quantification analyses.
    
\end{itemize}

\section{Methods}\label{sec:methods}

\subsection{Additive semi-parametric hybrid model}
Hybrid models integrate diverse sources of knowledge to create detailed representations. In this study, parametric physics-based and non-parametric data-driven submodels are combined to construct the regression function. 

On the one hand, parametric models assume a specific distribution of the output variables in relation to the input variables, with this distribution being characterized by a finite set of parameters \cite{mahmoud2019parametric}. Selecting an appropriate function requires prior knowledge, taking into account the expected relationship between the input and output variables. In the engineering field, a suitable choice may involve the underlying physical law that governs the problem. Parametric physics-based models may offer lower accuracy compared to non-parametric models; however, they are more interpretable, allowing for the identification of characteristic coefficients and providing a deeper understanding of the relationship between the modeled variables. While still offering a reasonable approximation, the error in a parametric regression may include noise, outliers, deviations from the expected behavior, or unmodeled influences from other observable variables. 

On the other hand, non-parametric models do not assume any specific form on the function used to calculate the dependent variables. Non-parametric models, such as neural networks, can effectively capture complex relationships among the considered variables. These models are particularly useful when the physical laws governing the system are unknown or too intricate, as they do not require domain-specific expertise to define the regression function. Additionally, they offer the flexibility to incorporate extra input variables into the regression task as needed. While they can achieve high accuracy when sufficient data are available, non-parametric models tend to be less interpretable and may encounter robustness issues when generalized to different conditions.

In real-world problems, typically, some aspects of the system can be described using a physical model, while other parts are more complex, and no effective parametric functions are available. The approximation provided by the physical model can be enhanced by modeling the residuals with a non-parametric model. In this paper, an additive semi-parametric model is proposed, to address the limitations of the two previously mentioned approaches:
\begin{equation}
         {\hat{y}} = f_{\text{phys}}(\mathbf{x}) + f_{\text{res}}(\mathbf{\Tilde{x}}) \, .
\end{equation}
The prediction consists of a physics-inspired part and a non-parametric part, the latter is specifically designed to predict the residuals of the physics-inspired output with respect to the global target. The physics-inspired submodel is driven by input variables $\mathbf{x}$ that are readily interpretable, whereas the residual submodel can accommodate a broader set of input variables $\mathbf{\Tilde{x}}$. These additional variables include both those that are easily interpretable and those that are more difficult to integrate into physical equations. The training of the hybrid model is performed in two steps. First, the physics-inspired submodel is fitted using the global target $y$ as the objective, providing a robust approximation of this quantity. Second, the residual submodel is trained to approximate $r = y-f_{\text{phys}}(\mathbf{x})$, enhancing the prediction by incorporating corrections based on unknown physical factors and integrating variables contained in $\mathbf{\Tilde{x}}$ that more effectively describe the state of the system. 

\subsection{Explainability and Uncertainty Quantification}

Once the hybrid model is described, two approaches are proposed to enhance the interpretation and reliability of the predictions obtained: SHAP (SHapley Additive exPlanations) and uncertainty quantification.

SHAP \cite{shap2017} is a game theory-based technique used for  measuring the marginal contribution of different input features of a model to its output. Its application involves the calculation of SHAP values, which quantify the local contribution $\phi_i$ of each feature $i$ to the output of the model $f$, such that:

\begin{equation}
    \phi_i = \sum_{\mathcal S \subseteq \mathcal N \\ - \{i\}} \frac{|\mathcal S|! \cdot (|\mathcal N| - |\mathcal S| -1)!}{|\mathcal N|!} [f(\mathcal S \cup \{i\}) - f(\mathcal S)]
    \,,
\end{equation}

where $\mathcal N$ is the total set of features, and $\mathcal S$ is any subset of features that does not include $i$. Thus, $f(\mathcal S \cup \{i\})$ represents all the possible outputs of the model $f$ when all the features are considered, while $f(\mathcal S)$ corresponds to the possible outputs of the model when the feature $i$ is ignored.

A relevant property of SHAP values is additivity, meaning that if a model can be decomposed into the sum of two submodels, the SHAP values can similarly be decomposed in an equivalent manner. SHAP also provides guarantees of consistency ---i.e., it is model-agnostic---, as well as robustness in the presence of missing features. Once the SHAP values are computed, the prediction for an instance $j$ can be reconstructed as the sum of the mean value of the predictive function over all instances, plus the sum of all the SHAP values:

\begin{equation}
    f(\mathbf{x}_j) = \bar{f} + \sum_{i\in\mathcal N} \phi_i(\mathbf{x}_j) \,.
\end{equation}

Furthermore, we employ Conformal Prediction (CP) \cite{angelopoulos2023conformal} to quantify the uncertainty of model outputs. In addition to predicted output values, CP provides confidence intervals (for regression tasks) or sets (for classification tasks) within which the true value will be contained, with at least a probability of $1 - \alpha$, where $\alpha$ determines the coverage.

To apply CP in regression tasks, we start with a pre-trained model and a calibration dataset.  Using the predictions on the calibration set, we compute a nonconformity score, which quantifies the difference $|\hat{y} - y|$ between model predictions $\hat{y}$ and true values $y$. Finally, we determine the $1 - \alpha$ percentile of these differences and generate an interval around the new prediction such that $[\hat{y} - \epsilon, \hat{y} + \epsilon ]$, where $\epsilon$ is the percentile considered.

In our case, we use the Conformalized Quantile Regression (CQR) variant proposed in \cite{romano2019conformalized}, which is available in the MAPIE library \cite{mapie2023}. This approach extends standard CP by using specific models ---or biased versions of the base model--- to determine interval bounds. For a given value of $\alpha$, we consider: (i) the base model, (ii) a model that predicts the lower bound of the intervals (with $\alpha / 2$), and (iii) a model that predicts the upper bound of the intervals (with $1 - \alpha / 2$). To obtain the interval models, we use the pinball loss function, which biases the base model predictions towards upper and lower bounds.

\section{Case study and experimental setup}\label{sec:case-study}

\subsection{Wind turbines physical background}
Although challenging, the modeling of the low-scale aerodynamic behavior of WTs can be accomplished using physics-based fluid dynamics methodologies. However, the predictive capacity of these models for the power generation of utility-scale wind farms is limited \cite{Howland2019}. Alternatively, partial physical insights can be obtained through well-established equations that relate certain large-scale variables. The power extracted by a WT from the kinetic energy of the incoming wind is expressed as:
\begin{equation}\label{eq:Power-cp}
    P = \frac{1}{2} C_{p} \rho A v^3\,,
\end{equation}
where $C_p$ is the power coefficient, $\rho$ is the air density, $A$ is the area swept by the blades of the WT, and $v$ is the wind velocity.

The power generated by a WT is closely related to the power coefficient $C_p$, a dimensionless parameter influenced by the turbine's intrinsic characteristics, such as its size, geometry, and aerodynamic properties, as well as operational conditions defined by variables like wind speed and pitch angle. A typical objective is to optimize $C_p$ to achieve maximum efficiency in converting wind kinetic energy into electrical energy, constrained by the theoretical upper limit of 0.5926, known as the Betz limit \cite{Betz1}. In regions with variable wind speeds, optimal power output is achieved by precisely adjusting the pitch angle, $\theta$, which represents the angle between the lateral axis of the blades and the direction of the relative wind. However, the complexity of pitch control arises from the nonlinear dynamics inherent in these systems, as well as from external disturbances.

The power coefficient's functional form is not dictated by physical laws and is often approximated using empirical parametric functions. To improve flexibility, we employ more adaptable regressors, such as neural networks. The output power is then computed using \autoref{eq:Power-cp} and depicted in \autoref{fig:model_diagram}, as further detailed in \autoref{subsec:hybrid_moel}. 

\subsection{Data description and preprocessing}

\begin{table}[t]
\centering
\footnotesize
\begin{tabular}{clc}
\hline
\textbf{Symbol} & \textbf{Name}       & \textbf{Unit}       \\           \hline
$v$              & Wind speed                  & m/s                      \\ 
$\theta$         & Pitch angle                     & rad                   \\   
$\omega$         & Rotor angular speed          & rad/s                    \\ 
$T_\text{out}$              & Outdoor temperature                    & $^\circ$C  \\ 
$T_n$              & Nacelle temperature                    & $^\circ$C           \\ 
$T_\text{r}$              & Rotor temperature                    & $^\circ$C      \\
$\alpha_\text{v}$              & Vane angle                    & rad       \\ 
$\alpha_\text{w}$              & Wind direction                    & rad   \\ 
$P$              & Generated power                    & kW                  \\ \hline
\end{tabular}
\caption{Symbols, names, and units of the primary features characterizing the state of a wind turbine}
\label{tab:features}
\end{table}

Real data were collected from two periods, 2013–2016 and 2017–2020, for four MM82 turbines manufactured by Senvion at the `La Haute Borne' wind farm \cite{LaHauteBorne1,LaHauteBorne2}, sourced from \href{https://www.engie.com/en}{ENGIE Renewables}. The original dataset comprises 10-minute averages of over 35 features related to turbine performance and weather conditions. The most important features are presented in \autoref{tab:features}.

\begin{figure}[t]
    \centering
    \includegraphics[width=1.0\textwidth]{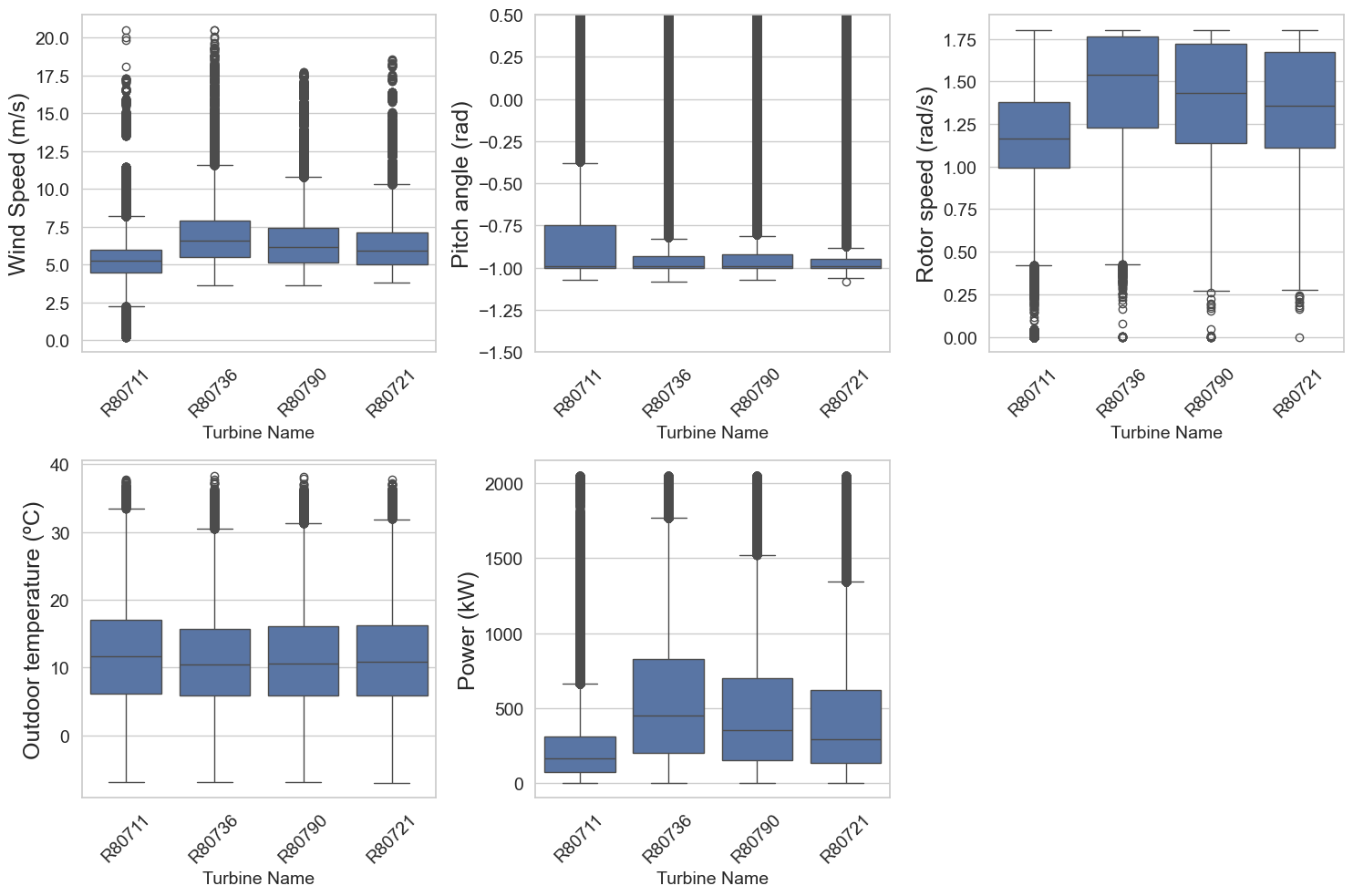}
    \caption{Box plots of wind speed, pitch angle, rotor speed, outdoor temperature and power for the four different wind turbines.}
    \label{fig:Vdistro}
\end{figure}

\begin{figure}[h!]
    \centering
    \includegraphics[width=0.7\textwidth]{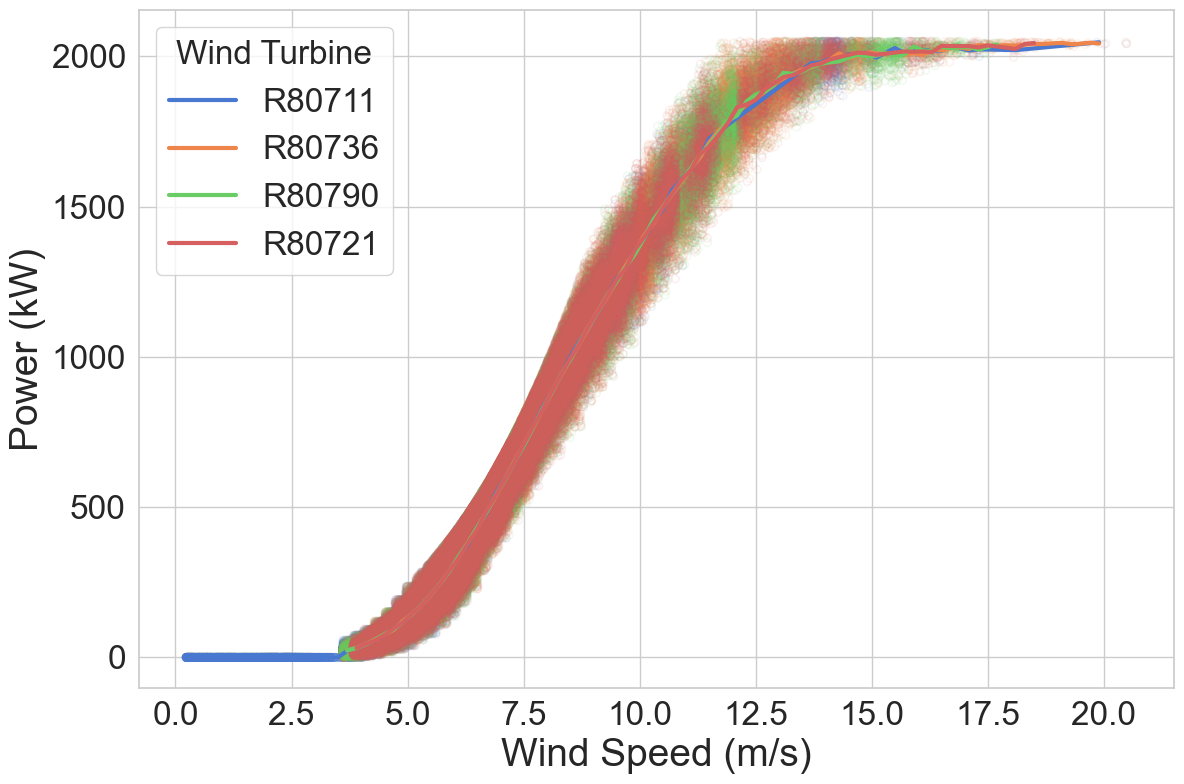}
    \caption{Average power curve derived from the data of the four wind turbines.}
    \label{fig:Pcurve_turbines}
\end{figure}

The wind farm under study consists of four turbines with similar data distributions, as illustrated in \autoref{fig:Vdistro}. Furthermore, the turbines share an identical power curve, shown in \autoref{fig:Pcurve_turbines}. Consequently, a single model is proposed to represent all the data without distinguishing between individual turbines.

Before preprocessing, the dataset comprised approximately 1 million instances. Initially, non-physical data points, such as those with a power coefficient exceeding the theoretical Betz limit ($C_p>0.5926$), were removed. Since the calculation of $C_p$ from direct measurements is sensitive to error propagation, only physically valid values were retained. Anomalous data were then identified by comparing the measured power with an estimate obtained from the power curve using an iterative median technique. Data points deviating from the median by more than three standard deviations (3$\sigma$) were classified as anomalies and excluded. Furthermore, a low-velocity power cutoff was applied to eliminate noisy data at low wind speeds, where the relative error has a more pronounced impact. As a result of these steps, the dataset was reduced from 1 million instances to approximately $7\times10^5$, representing 70\% of the original data.

\subsection{Power hybrid model}\label{subsec:hybrid_moel}

\begin{figure}[t]
    \centering
    \includegraphics[width=1.0\textwidth]{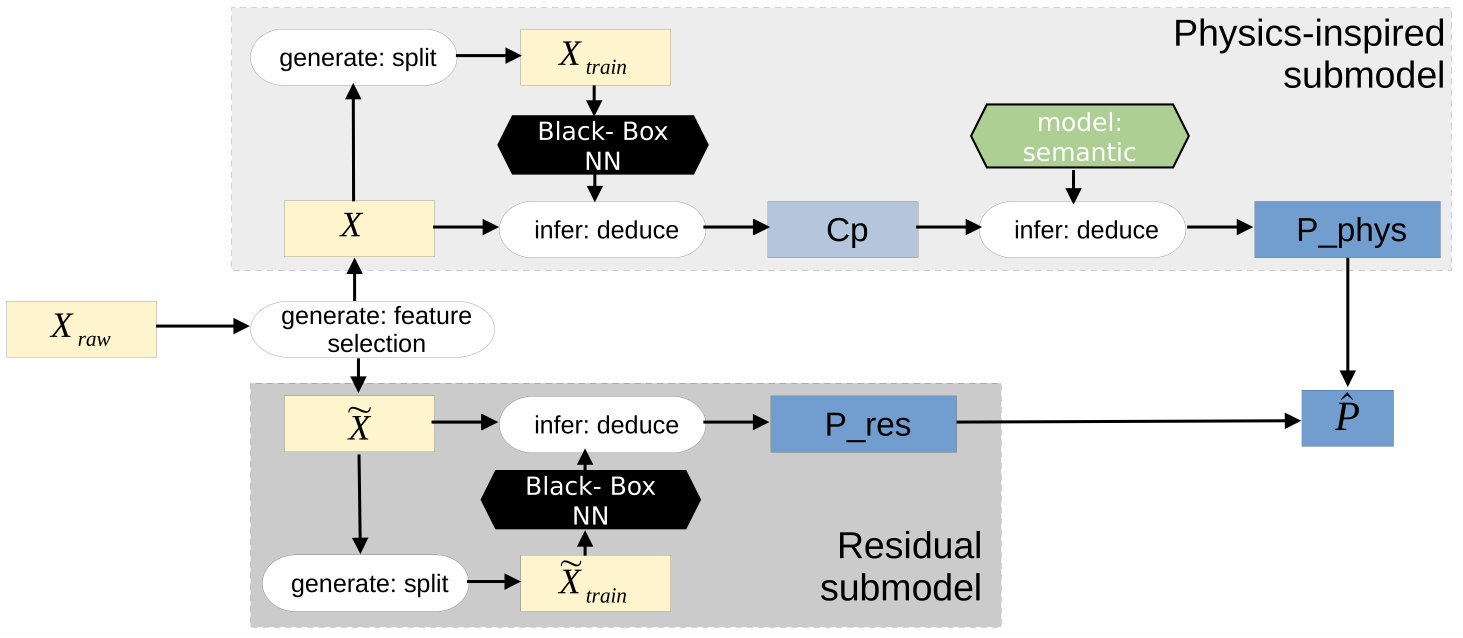}
    \caption{Diagram of the hybrid model, according to the taxonomy presented in \cite{vanBekkum2021}. Rectangular, rounded, and hexagonal boxes represent data, functions, and models, respectively. Yellow and blue boxes are used for inputs and outputs, respectively.}
    \label{fig:model_diagram}
\end{figure} 

For our purposes, the power prediction consists of two primary components: a physics-inspired part, based on \autoref{eq:Power-cp}, and a non-parametric part designed to predict the residuals of the physics-inspired output relative to the target variable, as shown in \autoref{fig:model_diagram}.
\begin{equation}\label{eq:hybrid_model}
     \hat{P} = P_{\text{phys}}(\mathbf{x}) + P_{\text{res}}(\mathbf{\Tilde{x}})
 \end{equation}
 The physics-inspired submodel $P_\text{phys}$ is driven by input variables directly related to the kinetic-electrical energy conversion process, such us wind velocity, pitch angle and rotor angular velocity. Additionally, the residual submodel $P_\text{res}$ uses a broader set of input variables extracted from the dataset, difficult to interpret and incorporate into physical equations, including outdoor temperature, nacelle temperature, rotor temperature, vane angle and wind direction:
\begin{gather}
    \mathbf{x} = (v,\theta,\omega) \,, \\
    \mathbf{\Tilde{x}} = (v,\theta,\omega,T_{\text{out}},
T_{\text{n}},T_{\text{r}},\alpha_\text{v},\alpha_\text{w}) \,.
\end{gather}
It should be noted that we decided to include the physically modeled variables $(v, \theta, \omega)$ in the residual non-parametric model to correct potential deviations of the data from \autoref{eq:Power-cp}, which is an approximation derived from physical laws. The physical submodel, in turn, incorporates an undetermined element: the power coefficient, whose functional form is not defined by physical laws. To address this, we employ a neural network for the regression of $C_p$, with a sigmoid output layer to constrain the predicted power coefficient to values within the range set by the Betz limit.

Although the variables $(v, \theta, \omega)$ are inputs to both the physics-based and residual submodels, for practical purposes of conducting the explainability analysis of the hybrid model, they are treated as distinct variables within the two submodels. This allows the physics-based and residual components to be reconstructed from the SHAP values of the hybrid model as:   
\begin{align}
    P_{\text{phys}} &\simeq \bar{P}_{\text{phys}} + \sum_{i=1}^3 \phi_i \,, \\
    P_{\text{res}} &\simeq \bar{P}_{\text{res}} + \sum_{i=1}^8 \Tilde{\phi_i} \,,
\end{align}
where $\phi_i$ and $\Tilde{\phi_i}$ are the SHAP values of features $\mathbf{x}$ and $\mathbf{\Tilde{x}}$, respectively.

\subsection{Computational details}
Our experimental setup is based on artificial neural network models, trained using the \href{https://www.tensorflow.org/}{\emph{Tensorflow}} library and hyperparameters optimized with the \textit{hyperband} algorithm from the \href{https://keras.io/keras_tuner/}{\emph{Keras Tuner}} library. All the calculations were carried out on a computer equipped with an 11th Gen Intel Core i7-11800H processor, 16 GB RAM memory and NVIDIA GeForce RTX 3060 graphics card.

Before training, a grid search was conducted to optimize hyperparameters, including the activation function, learning rate, number of hidden layers, and neurons per layer. The network was then trained using the optimal hyperparameters, and its performance was evaluated based on the epoch with the minimum loss value. Both submodels were trained using a batch size of 128, with the mean absolute error as the loss function. The dataset was randomly split into 80\% for training and 20\% for testing. An adaptive learning rate strategy was employed to reduce the learning rate when the loss function stabilized, with training continuing until the loss plateaued, typically after approximately 150 epochs.

The code used for data preprocessing, building, training, and evaluating the models is available at the following Github repository:\\ \href{https://github.com/alfonsogijon/WindTurbines_hybrid}{https://github.com/alfonsogijon/WindTurbines\_hybrid}.

\section{Results and discussion}\label{sec:results}

\subsection{Regression analysis}
Before initiating the model fitting process, a cross-correlation analysis was conducted to evaluate the influence of input variables on the output. As anticipated, variables utilized in the physical model, such as wind speed and rotor speed, exhibited strong correlations with the output, namely the power generated by the turbine. This finding reinforces the validity of the physical model using data-driven evidence. By setting a threshold of 0.9 for the correlation coefficient, the most critical variables representing the system's state were identified, while less relevant and redundant variables were excluded. This process reduced the original dataset from 35 variables to 9, as detailed in \autoref{tab:features}. It was observed that including additional variables did not improve accuracy significantly, while hindering the interpretation of the models.

\begin{table}[t]
\centering
\begin{tabular}{|c|c|c|c|c|}
\cline{2-5} \cline{3-5} \cline{4-5} \cline{5-5} 
\multicolumn{1}{c|}{} & Physics-based & Data-based & Hybrid \tabularnewline
\hline 
MAE (kW) & 16.31 & 11.97 & \textbf{11.73} \tabularnewline
\hline 
RMSE (kW) & 30.58 & 25.04 & \textbf{24.48} \tabularnewline
\hline 
MAPE (\%) & 3.71 & 2.35 & \textbf{2.32} \tabularnewline
\hline 
R2 score & 0.9953 & 0.9969 & \textbf{0.9976} \tabularnewline
\hline 
\end{tabular}
\caption{Comparison of the performance metrics for physics-based, data-based and hybrid models of the generated power.}
\label{tab:metrics}
\end{table}

Using \autoref{eq:Power-cp}, a physics-based submodel $P_\text{phys}$ was trained to predict the power output. The trained model was then used to compute prediction errors, which were subsequently utilized to train the non-parametric residual submodel $P_\text{res}$. The hybrid model was constructed by adding the outputs of both submodels. \autoref{tab:metrics} presents the performance metrics of the proposed models on the test dataset, including a purely data-driven model that utilizes the same set of eight input features employed by the hybrid model. As can be observed,  the hybrid model outperforms the physics-based model, achieving a 28\% reduction in Mean Absolute Error (MAE) and a 37\% reduction in Mean Absolute Percentage Error (MAPE). In turn, the performance of the resulting hybrid model essentially matches that of the reference data-based black-box approach. In this manner, the hybrid model effectively combines the accuracy of data-driven methods with the interpretability of physics-based approaches, as will be demonstrated in the following.

\begin{figure}[t]
    \centering
    \includegraphics[width=0.8\textwidth]{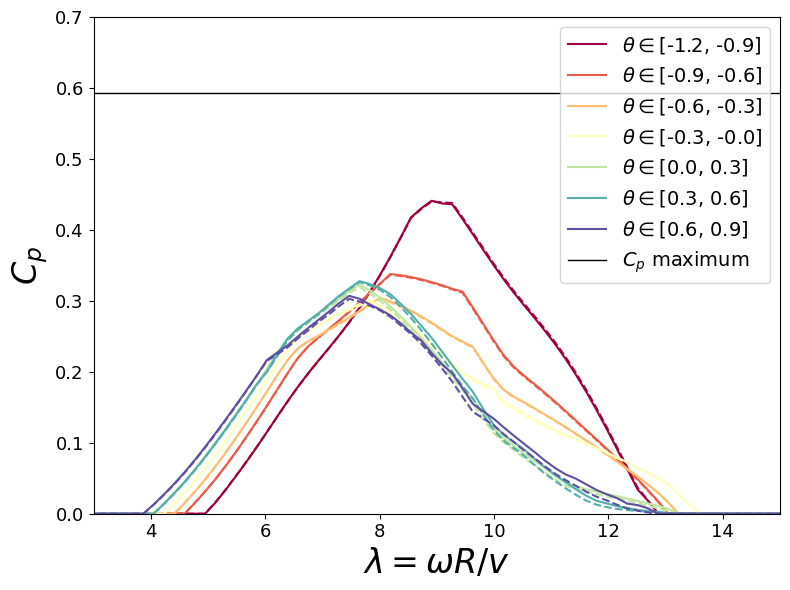}
    \caption{The $C_p$ curves derived from the physics-based model are represented as a function of the tip speed ratio $\lambda$, for various pitch angles in different colors (measured in radians). The observed data are represented with solid lines, while the model predictions are shown with dashed lines.}
    \label{fig:Cp_curves}
\end{figure}

The physics-based model provides a reliable approximation of the data, having a MAE of 16$\,$kW and a MAPE of $3.7$\%, as indicated by the metrics in \autoref{tab:metrics}. Additionally, it enables the identification and visualization of key parameters, such as the power coefficient $C_p$ of the WT. This coefficient can be expressed as a function of the tip-speed ratio, $\lambda=\omega R/v$, where $R$ represents the rotor radius, to derive a profile that characterizes the aerodynamic behavior of the WT. \autoref{fig:Cp_curves} illustrates the $C_p$ curve for different pitch angles, showing good agreement between the data and predictions, with the maximum $C_p$ shifting to the left as the pitch angle increases.

\begin{figure}[t]
    \centering
    \includegraphics[width=1.0\textwidth]{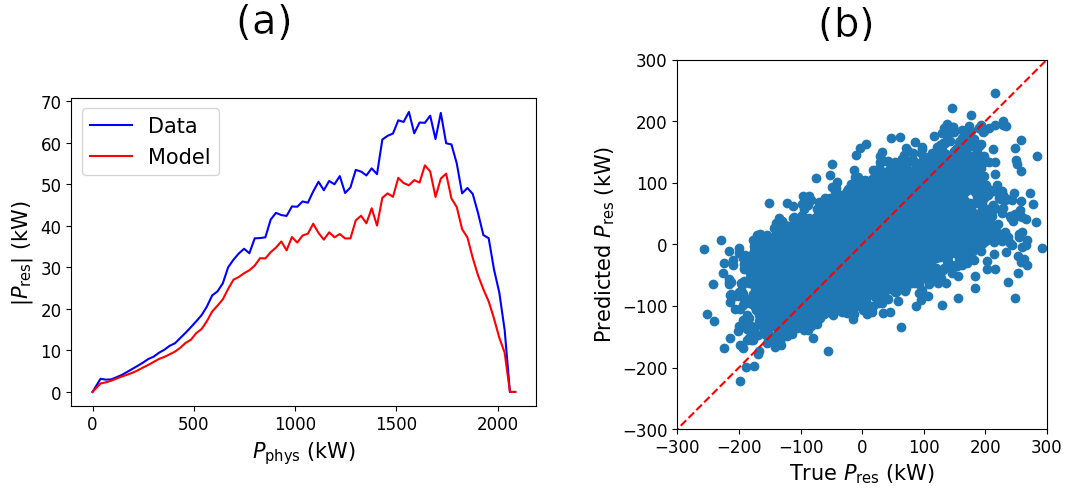}
    \caption{(a) Absolute residual between the true power and the prediction made by the physical model. (b) Predicted vs true residual power. \textcolor{red}}
    \label{fig:Pres}
\end{figure}

\begin{figure}[t]
    \centering
    \includegraphics[width=0.6\textwidth]{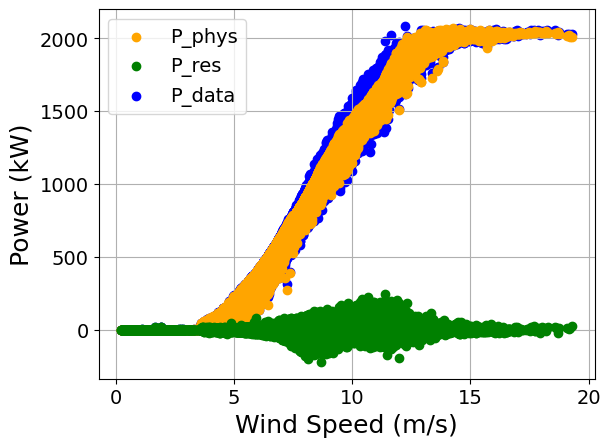}
    \caption{The contributions of the physics-based model (orange) and the residual model (green) to the predicted power are plotted as functions of wind speed, alongside the observed data (blue).}
    \label{fig:Pphys_Pres}
\end{figure}

The residual component of the hybrid model effectively captures the error between the target power and the prediction made by the physics-based submodel, as shown in \autoref{fig:Pres}. The panel (a) of the figure indicates that the absolute residuals are higher in the mid to high power range, suggesting that this region is not perfectly modeled by the physics submodel. Both the physics and residual components of the hybrid model are displayed in the scatter plot of \autoref{fig:Pphys_Pres}, alongside the test data. It is evident that the physics-based submodel (in orange) closely matches the power curve of the data (in blue), while the residuals are more pronounced at intermediate wind velocities, between 7.5 and 12.5 m/s.

\subsection{Explainability and Uncertainty Quantification}
To further enhance the interpretation and reliability of the hybrid model's predictions, we apply explainability and uncertainty quantification techniques.

As noted in the previous section, the physics-based component is fully interpretable, as it relies on a physical equation and an intermediate variable constrained within a defined range. In contrast, the residual component acts as a black box, taking input variables and generating predictions without explicit interpretability. The variables $(v,\theta,\omega)$ are inputs of both components of the hybrid model. To assess their significance within each submodel, two different instances of these variables are analyzed for the explainability study.

\begin{figure}[t!]
    \centering
    \includegraphics[width=1.0\textwidth]{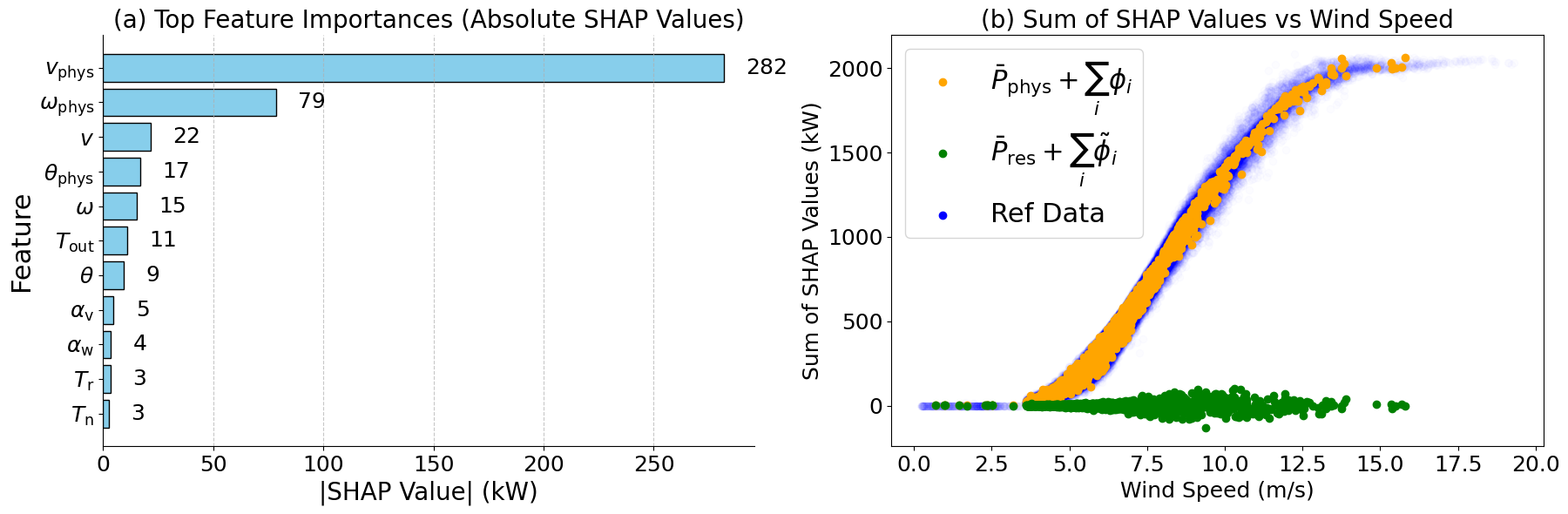}
    \caption{(a) Ranking of mean absolute SHAP values of the hybrid model. (b) Sum of SHAP values regarding physics and residual submodels.}
    \label{fig:SHAP_ranking_sum}
\end{figure}

The mean absolute SHAP values calculated over a sample of the test dataset and shown in panel (a) of \autoref{fig:SHAP_ranking_sum}, provide insights into the relative importance of each input feature on the output. As expected, the variables incorporated into the physics-inspired model, $(v_\text{phys},\theta_\text{phys},\omega_\text{phys})$, are the most influential. However, the corresponding instances of these variables in the residual model also exhibit high SHAP values, ranging from 10 to 20 kW. This indicates that the dependence of power on these variables is not fully captured by \autoref{eq:Power-cp}. Furthermore, the magnitude of the mean SHAP value associated with outdoor temperature indicates a notable influence on the predicted power, making it a good candidate for inclusion in a more sophisticated and accurate physical model.

An interesting property of the additive hybrid model is its ability to recover the contributions of its two components by summing the SHAP values associated with each submodel. When the average power plus the sum of the SHAP values is plotted against wind speed, the sigmoid shape characteristic of the physics-based submodel emerges, while the residual component oscillates around zero, exhibiting larger amplitudes at intermediate wind velocities, as shown in panel (b) of \autoref{fig:SHAP_ranking_sum}.

\begin{figure}[htbp]
    \centering
    \includegraphics[width=1.0\textwidth]{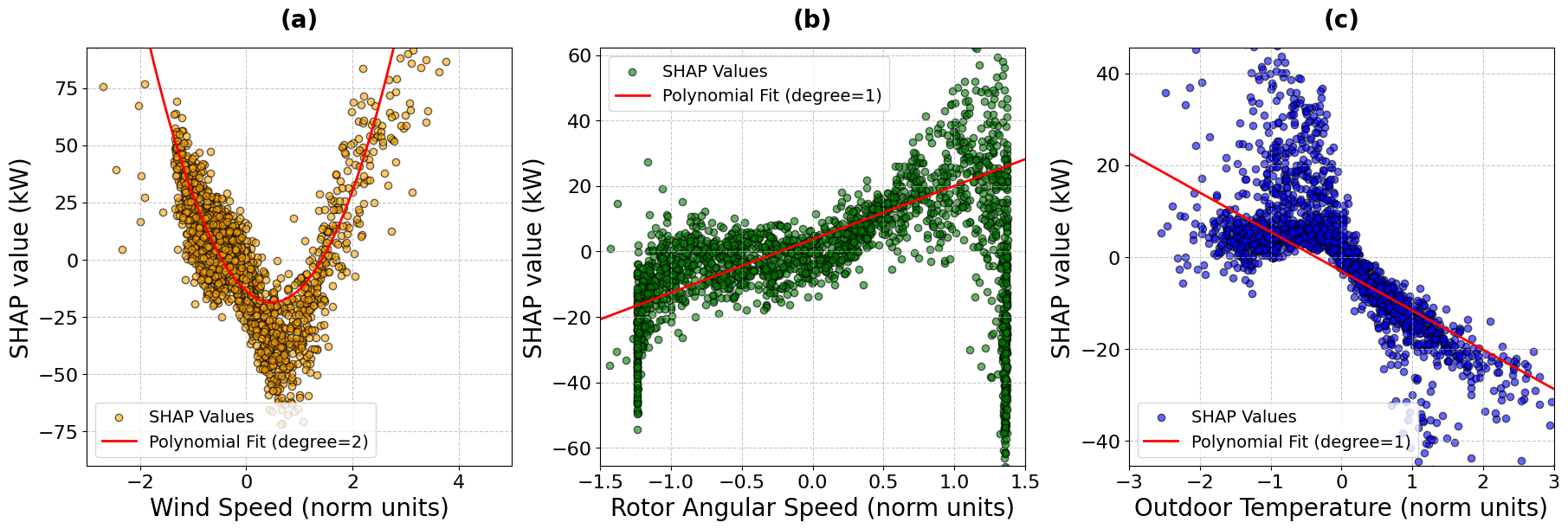}
    \caption{SHAP value of the residual power model vs. associated variable for (a) wind speed,(b) rotor speed and (c) outdoor temperature. Linear and quadratic regressions are shown in red color.}
    \label{fig:SHAP_correlations}
\end{figure}

At this stage, it is natural to question whether the hybrid model can provide insights for developing new, more sophisticated physics-based models. To delve deeper into the contribution of input variables to the residual model's output, scatter plots of SHAP values against their corresponding variables can be analyzed. \autoref{fig:SHAP_correlations} illustrates the influence of the most significant variables, accompanied by low-order polynomial regression fits to capture underlying trends. 
The positive correlation between the SHAP value and rotor angular speed indicates that this variable positively contributes to the power residual, suggesting that the physical model underestimates its impact. Conversely, the negative correlation between the SHAP value and outdoor temperature suggests that an increase in temperature should decrease the power output of the physical model. This is logical, as higher temperatures reduce the efficiency of electrical energy conversion, a factor not accounted for by the physical model. Lastly, the correlation between the SHAP value and wind speed implies that the power output of the physical model should be adjusted for both higher and lower velocities than the mean, as the SHAP value is positive in both directions.
However, the observed correlations between SHAP values and input variables do not necessarily imply a causal relationship between those input variables and the generated power \cite{shap_causal_insights}. In fact, attempts to approximate the residual power by the sum of polynomials of the input variables, as suggested by \autoref{fig:SHAP_correlations}, result in poor regression performance. This is mainly due to the interdependence of input variables and the complex correlations among them. In fact, to fully analyze the dependence of the target variable $P$ on the input variables, it is necessary to consider the sum of all SHAP values, as demonstrated in the bottom panel of \autoref{fig:SHAP_ranking_sum}. Tools like SHAP are effective in identifying the most informative relationships between input features and the predicted outcome, providing valuable insights into the model's behavior. However, to develop causal or physical models, it is necessary to make assumptions and leverage the methodologies of causal analysis. 

\begin{figure}[t]
    \centering
    \includegraphics[width=0.8\textwidth]{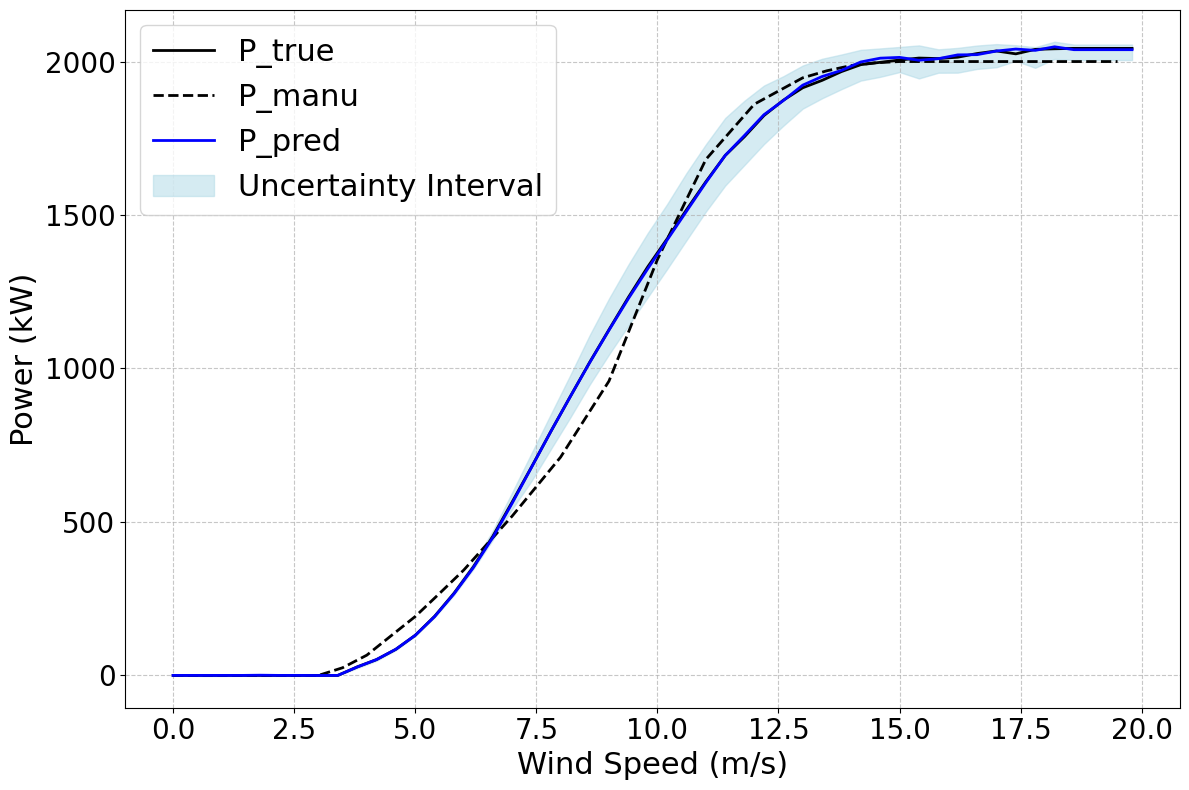}
    \caption{Power curve comparing model predictions with uncertainty intervals (blue), data (solid black line) and manufacturer specifications (dash black line). The data curve is nearly indistinguishable to the human eye, as it closely overlaps with the model curve, lying beneath it.}
    \label{fig:PvsV_uncertainty}
\end{figure}

Finally, to evaluate the reliability of the predictions, uncertainties are calibrated using a calibration set comprising half of the test set. The model's performance is then assessed using prediction intervals on the remaining portion of the test set. This procedure does not require re-training the hybrid model, but instead involves training the upper (95\%) and lower (5\%) quantile estimations of the predictions. The conformalized quantile regression method is then used to provide conformal predictions, generating target predictions for new samples with associated confidence intervals. The effective coverage is determined by estimating the fraction of true labels that fall within the prediction intervals. In this case, the coverage is 86\%, and the prediction intervals have a mean length of 49\,kW.

The predicted power curve with uncertainty estimation is shown in \autoref{fig:PvsV_uncertainty}, alongside the mean curve derived from the test data and the manufacturer's specifications. As observed, the uncertainty intervals are negligible at low wind speeds and become wider at mid to high speeds, providing an indication of the regions where the power model is more or less accurate. Furthermore, these results are consistent with previous research \cite{Gijon2023}, which found larger uncertainties in the same wind speed range when using physics-informed neural network models. Discrepancies with the theoretical manufacturer curve are typical in large datasets from real wind farms, and often does not accurately reflect the actual performance of WTs \cite{WT_power_curve,PAGNINI2015,YAN2019}.

\section{Conclusions}\label{sec:conclusions}
In this study, we design and validate a hybrid semi-parametric model comprising a physics-inspired submodel, $P_\text{phys}$ and a non-parametric submodel for predicting the residual power of the physical term, denoted as $P_\text{res}$. The proposed model, trained on real historical data from four turbines at `La Haute Borne' wind farm, demonstrates an improvement of a 37\% in predicting generated power.

The physics-inspired submodel offers inherent interpretability of the power coefficient, as it is built upon a physical equation that captures the relationships among the system’s most critical variables. In contrast, the non-parametric residual submodel, which accounts for the unexplained components of the system, requires an additional analysis to interpret its behavior, which is accomplished using the SHAP explainability technique. Moreover, the conformalized quantile regression method provides conformal predictions along with associated confidence intervals, consistent with the data and previous results. The integration of these components results in a flexible, accurate and reliable framework.

Once deployed, this hybrid model can serve as a robust regression-based anomaly detection tool,  quantifying the probability of new data being classified as normal or anomalous. Furthermore, all models presented in this study are fully differentiable, making them suitable for developing optimal operation controllers. These controllers have the potential to further enhance power generation efficiency across varying wind speed conditions.

Using SHAP values, the model also provides insights into the relative importance of input features and their influence on the predicted power output. The analysis indicates that incorporating outdoor temperature into future, more advanced physical models could enhance prediction accuracy. Additionally, it highlights potential adjustments in the relationships between power and wind or rotor speed. Future research should focus on a more in-depth analysis of correlations and the application of causal analysis tools, which would be valuable for the development of improved physics-based causal models.

Although the hybrid model shows significant potential, further research is required to assess its robustness and scalability across different wind farm datasets. Nevertheless,  the model can be readily adapted to different manufacturers by fine-tuning the parameters of the pre-trained residual submodel and updating the turbine-specific parameters of the physics-based submodel. Furthermore, the proposed methodology is highly versatile and applicable to a wide range of scenarios where a physics-based model approximates key dynamics. By incorporating additional data through an additive non-parametric submodel, the framework effectively captures residual components and integrates previously unknown physical phenomena, all without the need to retrain the physics-based model.

\section*{CRediT authorship contribution statement}
\textbf{A. Gijón:} Conceptualization, Methodology, Software, Data Curation, Formal Analysis, Investigation, Validation, Visualization,  Writing - Original Draft, 
\textbf{S. Eiraudo:} Conceptualization, Methodology, Software, Investigation, Writing - Review \& Editing,
\textbf{A. Manjavacas:} Methodology, Software, Investigation, Writing - Review \& Editing, \textbf{D.S. Schiera} Supervision, Writing - Review \& Editing, 
\textbf{M. Molina-Solana:} Supervision, Funding acquisition, Project administration, Writing - Review \& Editing, \textbf{J. Gomez-Romero:} Supervision,  Funding acquisition, Project administration, Writing - Review \& Editing.

\section*{Declaration of competing interest}
The authors declare that they have no known competing financial
interests or personal relationships that could have appeared to influence
the work reported in this paper.

\section*{Acknowledgments}
This work was primarily funded by the Spanish Ministry of Economic Affairs and Digital Transformation (NextGenerationEU funds) within the project IA4TES MIA.2021.M04.0008. It was also partially funded by ERDF/Junta de Andalucía (D3S project P21.00247, and SE2021 UGR IFMIF-DONES), and MICIU/AEI/ \\10.13039/501100011033 and EU ERDF (SINERGY, PID2021.125537NA.I00).
The authors also acknowledge the ENGIE company for providing such an interesting and well-documented dataset.



\nocite{*}
\bibliographystyle{elsarticle-num} 
\bibliography{references.bib}

\end{document}